%% file: main_IEEE.tex
\documentclass[conference]{IEEEtran}
\pdfoutput=1
\IEEEoverridecommandlockouts
\usepackage{cite}
\usepackage{amsmath,amssymb,amsfonts}
\usepackage{algorithmic}
\usepackage{graphicx}
\usepackage{textcomp}
\usepackage{xcolor}
\usepackage{multirow}
\usepackage{bm}
\usepackage{makecell}
\usepackage[breaklinks=true,bookmarks=false,linkcolor=blue,filecolor=blue,citecolor=black,urlcolor=cyan]{hyperref}
 \hypersetup{
     colorlinks=true,
     linkcolor=blue,
     filecolor=blue,
     citecolor=black,
     urlcolor=cyan,
     }

\def\BibTeX{{\rm B\kern-.05em{\sc i\kern-.025em b}\kern-.08em
    T\kern-.1667em\lower.7ex\hbox{E}\kern-.125emX}}
\begin{document}

\title{DINs: Deep Interactive Networks for Neurofibroma Segmentation in Neurofibromatosis Type 1 on Whole-Body MRI}
\author{Jian-Wei Zhang, Wei Chen, K. Ina Ly, Xubin Zhang, Fan Yan, Justin Jordan, Gordon Harris, Scott Plotkin, \\ Pengyi Hao, and Wenli Cai
\thanks{J.W. Zhang, X. Zhang and F. Yan are with the State Key Lab of CAD\&CG, Zhejiang University, Hangzhou, 310012, China.}
\thanks{W. Chen, corresponding author, is with The State Key Lab of CAD\&CG, Zhejiang University, Hangzhou, 310012, China. (e-mail: chenvis@zju.edu.cn)}
\thanks{K.I. Ly, J. Jordan and S. Plotkin are with the Cancer Center, Massachusetts General Hospital, Boston, MA 02114, USA}
\thanks{G. Harris is with the Department of Radiology, Massachusetts General Hospital, Boston, MA 02114, USA}
\thanks{P. Hao is with the School of Computer Science and Technology, Zhejiang University of Technology, Hangzhou, 310024, China}
\thanks{W. Cai, corresponding author, is with the Department of Radiology, Massachusetts General Hospital, Boston, MA 02114, USA (e-mail: cai.wenli@mgh.harvard.edu)}
\thanks{© 2021 IEEE. Personal use of this material is permitted. Permission from IEEE must be obtained for all other uses, in any current or future media, including reprinting/republishing this material for advertising or promotional purposes, creating new collective works, for resale or redistribution to servers or lists, or reuse of any copyrighted component of this work in other works.}
}

\maketitle
\def\etal{{\it et al.}}

\begin{abstract}
    Neurofibromatosis type 1 (NF1) is an autosomal dominant tumor predisposition syndrome that involves the central and peripheral nervous systems.
    Accurate detection and segmentation of neurofibromas are essential for assessing tumor burden and longitudinal tumor size changes. Automatic convolutional neural networks (CNNs) are sensitive and vulnerable as tumors' variable anatomical location and heterogeneous appearance on MRI. 
    In this study, we propose deep interactive networks (DINs) to address the above limitations. User interactions guide the model to recognize complicated tumors and quickly adapt to heterogeneous tumors. We introduce a simple but effective Exponential Distance Transform (ExpDT) that converts user interactions into guide maps regarded as the spatial and appearance prior. Comparing with popular Euclidean and geodesic distances, ExpDT is more robust to various image sizes, which reserves the distribution of interactive inputs. Furthermore, to enhance the tumor-related features, we design a deep interactive module to propagate the guides into deeper layers. 
    We train and evaluate DINs on three MRI data sets from NF1 patients. The experiment results yield significant improvements of 44\% and 14\% in DSC comparing with automated and other interactive methods, respectively. We also experimentally demonstrate the efficiency of DINs in reducing user burden when comparing with conventional interactive methods. The source code of our method is available at \url{https://github.com/Jarvis73/DINs}.
\end{abstract}


\input{subsections/1-Introduction}
\input{subsections/2-RelatedWorks}
\input{subsections/3-Methods}
\input{subsections/4-Experiments}
\input{subsections/5-Results}
\input{subsections/6-Conclusion}

\section*{Acknowledgment}

Wei Chen is supported by The National Key R\&D Program of China under grant No.2019YFB1404802 and National Natural Science Foundation of China (61772456). Wenli Cai is supported by Grant R42CA192600 and R42CA189637 from the National Institute of Health and Children Tumor Foundation. Pengyi Hao is supported by National Natural Science Foundation of China under grants No.61801428. Scott Plotkin received support from the Department of Defense (W81 XWH-06-1-0739) and philanthropic funds.

\bibliographystyle{IEEEtran}
\bibliography{refs}

\end{document}

%% file: subsections/1-Introduction.tex
\section{Introduction}
\label{sec1}

\IEEEPARstart{N}{eurofibromatosis} type-1 (NF1) is an autosomal dominant neurogenetic disorder characterized by the development of both benign and malignant tumors. The hallmark tumors are neurofibromas, which are histologically benign tumors that arise from the peripheral nerve sheath and involve any body part. Despite benign histology, they can cause significant morbidity due to compression and invasion of nerves and other vital anatomical organs. Neurofibromas can be located deep inside the body and, if asymptomatic, are usually detected by whole-body magnetic resonance imaging (WBMRI) using short tau inversion recovery (STIR) sequences. Based on tumor morphology on MRI, plexiform (invasive or involving multiple nerves) neurofibromas (PNFs) carry an increased risk for transformation into malignant peripheral nerve sheath tumors.

\begin{figure*}[!t]
\centering
\includegraphics[width=0.83\linewidth]{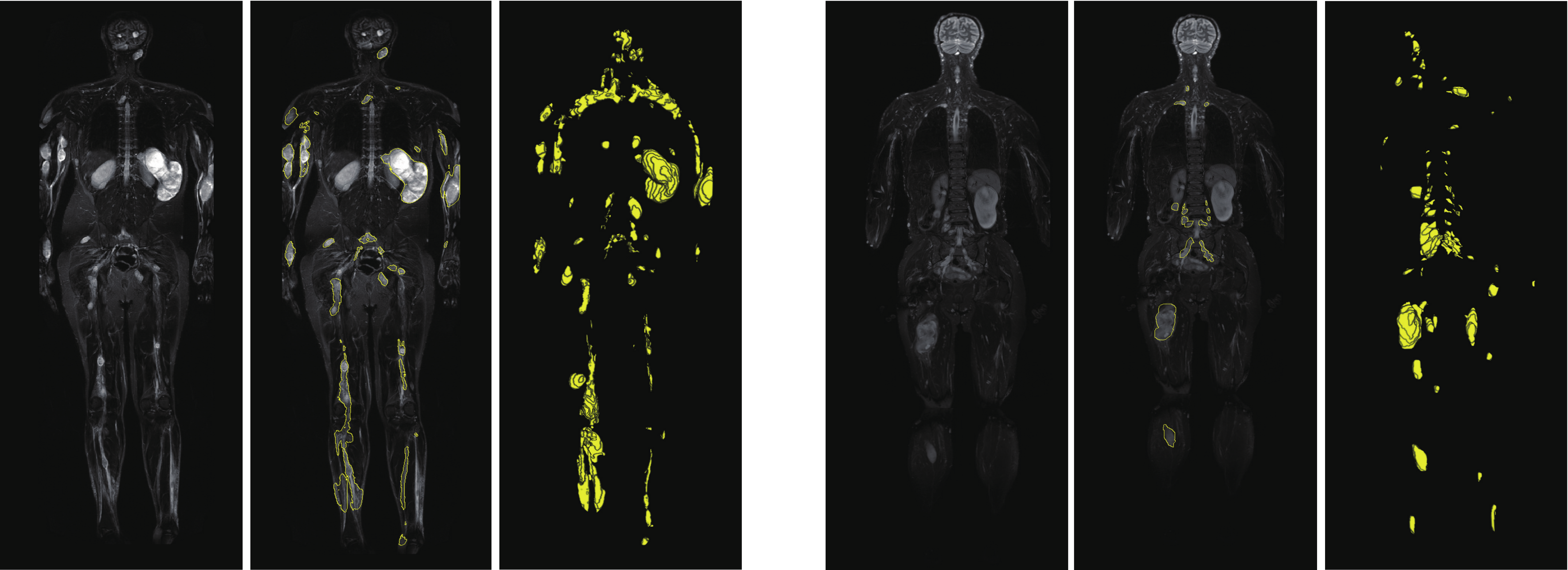}
\caption{Two examples of neurofibromas with ground truth on 3D WBMRI. The three images in each example are (1) a coronal slice, (2) annotations of neurofibromas contoured in yellow, (3) and the 3D view of neurofibromas, respectively.}
\label{fig:display}
\end{figure*}

Fig.~\ref{fig:display} depicts two NF1 cases of WBMRI with the ground truth segmentation of tumor regions contoured in yellow. Accurate detection and evaluation of tumor burden on WBMRI are important for longitudinal tracking of tumor size, which enables accurate assessment of tumor growth and treatment response. However, the detection and segmentation of neurofibroma on WBMRI, particularly PNFs, is associated with three technical challenges.

{\bf Large number of tumors across the entire body and variable anatomical locations of tumors.} Neurofibromas can develop anywhere along peripheral nerves. Their appearance across individuals can vary in number (from none to hundreds) and sizes (from several cubic centimeters to several thousand cubic centimeters). Traditional interactive segmentation methods cannot reasonably handle such a large number of tumors at a time. Conventional segmentation methods for neurofibromas have been proposed in the literature, including histogram thresholding~\cite{Weizman2014}, region growing~\cite{lior2012interactive} using histogram templates, 3D dynamic thresholding level-set~\cite{CAI2018144}. These interactive segmentation methods are labor-intensive and time-consuming since they involve manual identification of individual tumors by raters and interactive contour correction due to the imperfect segmentation obtained from automated methods. In general, it may take a few minutes to 1-2 hours to complete segmentation on WBMRI.

{\bf Heterogeneous and diffuse tumor architecture.} PNFs can be elongated in shape with a characteristic ringlike or septate pattern that typically has a target-like appearance on MRI, with central low signal intensity and peripheral high signal intensity. Recently, deep convolutional neural networks~(CNNs) have achieved great success in medical image segmentation, such as U-Net~\cite{unet}, V-Net~\cite{vnet}, DeepMedic~\cite{kamnitsas2017efficient}, and nnU-Net~\cite{nnunet}. CNNs have brought a breakthrough for tumor segmentation in the brain~\cite{HAVAEI201718}, lung~\cite{8417454}, liver~\cite{h-denseunet}, and other organs. Nevertheless, their application in neurofibroma segmentation on WBMRI has been minimal. Besides, CNN-based approaches tend not to generalize well to new data because the targeted neurofibromas may be substantially different in size, shape, intensity, and boundary to adjacent organs in the training and testing data sets. Here, we explore how to embed user interactions into CNNs to improve generalizability for obtaining an accurate and efficient interactive segmentation approach on WBMRI.

{\bf Guide maps suffer from the distribution shift for variable image sizes.} Some CNN-based interactive segmentation approaches~\cite{DIOS, DeepIGeoS} have been proposed to extract foreground objects interactively. These methods convert user interactions into distance maps utilizing either Euclidean distance transform~\cite{DIOS} (EDT) or geodesic distance transform~\cite{DeepIGeoS} (GDT). Typically, training CNNs with image patches and fine-tune/test on the whole image is a common trade-off between GPU memory and accuracy/inference speed~\cite{sekou2019patch,nazeri2018two,h-denseunet}. However, both transformations are sensitive to image sizes, which leads to the distribution shift for guide maps with various sizes and a performance decrease when applied to neurofibroma data. 

This paper proposes deep interactive neural networks (DINs) for interactive neurofibroma segmentation on WBMRI. We first adapt popular 3D U-Net~\cite{3d-unet} to the neurofibroma data on WBMRI by introducing anisotropic convolutional kernels for more accurate tumor-related feature extraction. Then, user interactions are encoded into guide maps as inputs with a distance transformation and embedded into multiple layers of the model for reserving user knowledge in deeper layers. The guide maps are regarded as the local appearance prior and the spatial prior. To avoid the effect of variable image sizes, we propose using exponential distance transform (ExpDT), whose intensity distribution is size-agnostic. With the guide maps, DINs embed users' prior knowledge into neural networks for correcting segmentation results. Furthermore, to reduce the interaction effort in training and testing processes of DINs, we develop a strategy to simulate user interactions to synthesize various user interactive methods and rapidly explore the best hyper-parameters. It is well known that medical images acquired from different devices have a distribution shift problem that can not be neglected and may lead to poor generalization for CNN models. In this situation, we experimentally demonstrated that DINs have stronger robustness and stability than previous automated and interactive methods.

To evaluate DINs, we collected two WBMRI data sets and a local-region MRI~(LRMRI) data set from NF1 patients, obtained using different MRI acquisition parameters. Experiments showed that DINs significantly outperformed automated methods by 44\% in Dice Similarity Score (DSC), which demonstrated the effectiveness of the CNN-based interactive segmentation. DINs outperformed other CNN-based interactive methods by 29\% and ExpDT outperformed other distance transforms (DTs) by 14\% in DSC. Furthermore, comparison results on conventional interactive methods suggested that DINs significantly reduced user interactions and running time. 

The main contributions of the work are summarized as follow:
\begin{itemize}
    \item We propose DINs to cope with the challenges of neurofibromas for interactive segmentation on WBMRI.
    \item We introduce ExpDT for integrating user interactions into neural networks. ExpDT is size-independent comparing with other common DTs and therefore is more suitable for WBMRI. 
    \item We propose a deep interactive module to integrate user knowledge into the deeper layers of the model, which effectively enhances the learned features about neurofibromas and improves the segmentation performance.
    \item We develop a strategy to simulate user interactions for training 3D interactive neural network models.
    \item DINs outperform automated and interactive methods by 44\% and 29\% in DSC, and ExpDT outperforms other DTs by 14\% in DSC. Furthermore, DINs reduce user interactions and running time comparing with conventional methods.
\end{itemize}

%% file: subsections/2-RelatedWorks.tex
\section{Related Work}

\subsection{Neurofibroma segmentation}
A few interactive and semi-automated segmentation methods have been developed for NF1 in the literature. Solomon \etal~\cite{solomon2004automated} developed an interactive 2D segmentation method for PNF that detects tumor regions within a manually defined area on each slice using a histogram-based threshold. This method failed if the histogram is unimodal or close to unimodal. Thus, manual contouring was frequently required to correct the resulting contours. Following this idea, Weizman \etal~\cite{Weizman2014} proposed 15 histogram templates of various distributions from bimodal to unimodal to identify the optimal threshold. Cai \etal~\cite{CAI2018144} developed the 3DQI system for semi-automated neurofibroma segmentation performed by the dynamic-threshold level set method starting from a seed region. These existing methods required a large amount of time and effort of interaction from users, either slice-by-slice or tumor-by-tumor with user-provided scribbles or initial seeds. However, as shown in Fig.~\ref{fig:display}, there might be dozens or hundreds of tumors in a single study, which put a heavy burden of interaction on users. Some work~\cite{WU202058,Ho2020} introduced neural networks into the segmentation of neurofibromas. Wu \etal~\cite{WU202058} integrated CNNs into the Active Contour Model for predicting the parametric maps, while Ho \etal~\cite{Ho2020} compared a multi-spectral neural network classifier with manual segmentation on diffusion-weighted imaging data. However, these methods still relied on conventional segmentation methods, and state-of-the-art deep CNNs were not explored for neurofibroma segmentation yet. Therefore, a highly accurate and efficient neurofibroma segmentation method remained a technical challenge.

\subsection{CNNs in medical image segmentation}
CNNs have been successfully adopted in various medical image segmentation applications. In particular, U-Net~\cite{unet} was designed for medical image semantic segmentation with the symmetric encoder and decoder paths. Long skip connections between the two paths enhanced the fusion of multi-level features. Based on the same network, many encoder-decoder CNNs have subsequently been introduced for 2D and 3D medical image segmentation. 
3D U-Net~\cite{3d-unet} expanded U-Net from 2D image segmentation to 3D volumetric image segmentation using 3D convolutions and a training strategy with sparse annotation. VNet~\cite{vnet} and HighRes3DNet~\cite{HighRes3DNet} incorporated residual modules~\cite{resnet} into their network structures. The difference between VNet and HighRes3DNet was that the VNet enlarged the receptive field by downsampling the feature map with large-stride convolution while HighRes3DNet adopted dilated convolution~\cite{deeplab}. In addition, cascaded networks~\cite{cascaded-fcn, h-denseunet}, were also explored for improving the learning ability of the model. H-DenseUNet~\cite{h-denseunet} adopted a multi-stage strategy, cascaded 2D and 3D networks to jointly fuse and optimize the learned intra-slice and inter-slice features for better liver and tumor segmentation.
These methods have achieved promising results in various applications of tumor and organ segmentation. However, existing CNN models did not generalize well to segment tumors in new data sets with different spatial and density distributions (such as variable location and tumor morphology in NF1) than in the training set.

\subsection{Interaction in CNNs}
Inspired by the semi-automated image segmentation techniques, which achieved a higher accuracy with minimal effort of interaction to provide cues that guide segmentation algorithms, such as clicks or scribbles used in Graph Cut~\cite{graphcut} and Random Walk~\cite{randomwalk}, recent works have introduced user interactions as extra channels of the input images in CNN~\cite{DIOS,DeepIGeoS}. The user interactions were typically transformed into guide maps by DT. We reviewed two well-known DTs:
\begin{itemize}
    \item {\it Euclidean distance transform.} Xu \etal~\cite{DIOS} proposed Deep Interactive Object Selection~(DIOS) for natural image segmentation, directly fine-tuned the fully convolutional network (FCN)~\cite{fcn} and introduced EDT for adapting interaction and refined the segmentation results by graph cuts. 
    \item {\it Geodesic distance transform.} Wang \etal~\cite{DeepIGeoS} proposed two networks for 2D placenta segmentation: a proposal network (P-Net) for initial segmentation, and a refinement network (R-Net) for refined segmentation. GDT was used to transform user interactions into intensity-related distance maps that could provide auxiliary information for accurate segmentation.
\end{itemize}
Both methods transformed the user clicks into guide maps using a DT. However, the intensity distribution of the two distance maps closely relied on the image size which was commonly variable in 3D medical image segmentation. Inconsistent intensity distribution of the input guide maps significantly affected the segmentation accuracy of CNN-based models. Instead of interacting in the training stage, image-specific fine-tuning~\cite{Image-Specific-Fine-Tuning} incorporating user interactions at the test stage was another solution. Nevertheless, fine-tuning a deep neural network at runtime required massive computational resources, which was unavailable in most of the situations. 

%% file: subsections/3-Methods.tex
\begin{figure*}[!t]
    \centering
    \includegraphics[width=\linewidth]{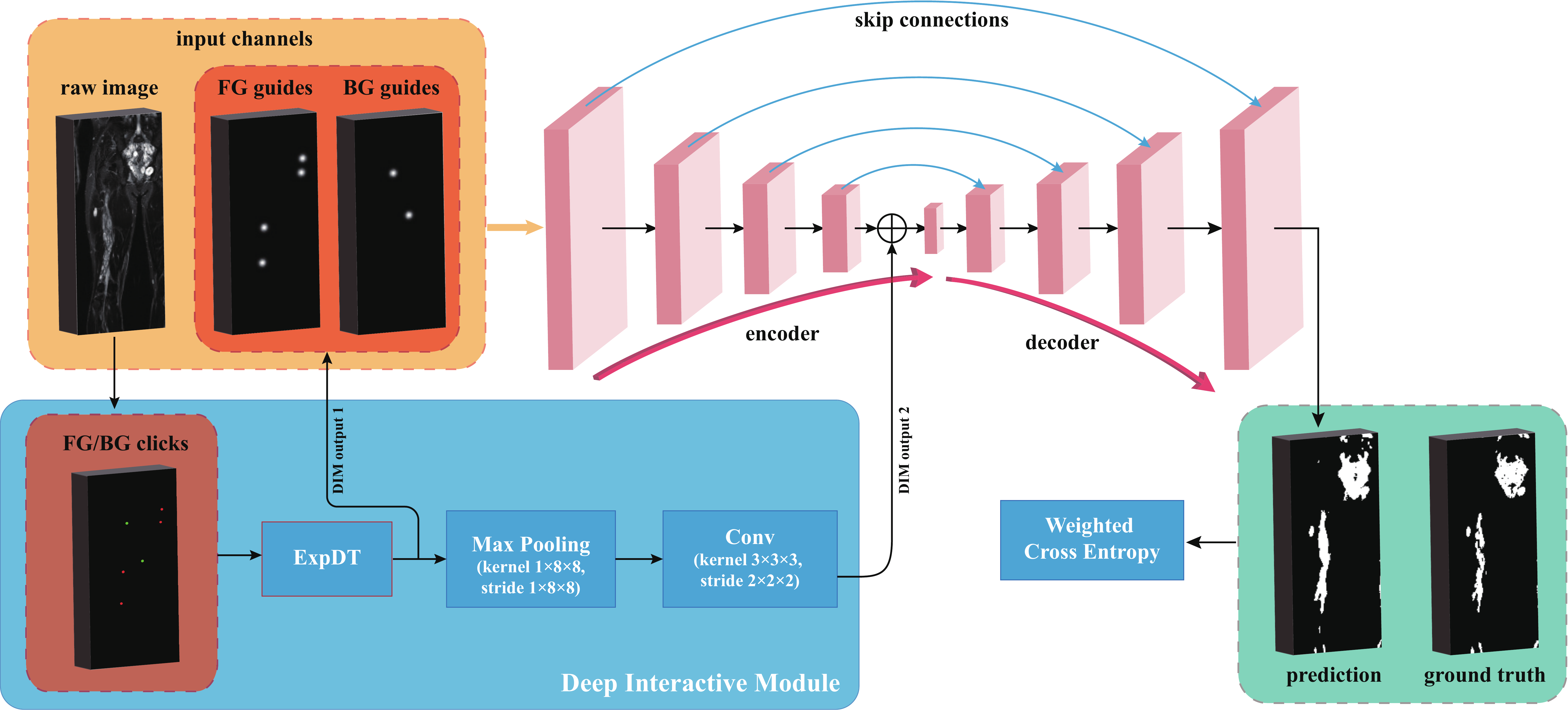}
    \caption{The structure of DINs. User clicks are transformed into foreground guides and background guides, which are concatenated with raw images as the input of backbone networks. ExpDT is the exponential distance transform. The two guides are also encoded and integrated into deeper layers according to the Deep Interactive Module for enhancing information flow about the interactions.}
    \label{fig:framework}
\end{figure*}

\section{Methods}
In the context of neurofibroma segmentation on WBMRI, we propose deep interactive neural networks~(DINs) for 3D medical image semantic segmentation, which is inspired by deep object selection~\cite{DIOS} for interactive segmentation of natural images and feature modulation~\cite{film, efficient} for conditional controlling of the neural networks. DINs employed an encoder-decoder backbone embedded within the deep interactive module~(DIM). The DIM influences neural network segmentation via the image-specific information generated from user interactions represented by the distance map. The structure of DINs is shown in Fig.~\ref{fig:framework}. Furthermore, for efficient training of DINs, we propose a strategy to simulate user interactions in the training process, thereby avoiding creating thousands of training samples manually. This strategy can also be used to evaluate the performance of DINs and adjust the hyper-parameters quickly. 

\subsection{Exponential distance transform}
\newcommand{\xx}{\mathbf{x}}
\newcommand{\yy}{\mathbf{y}}
\newcommand{\uu}{\mathbf{u}}
\newcommand{\Ga}{\bm{\Gamma}}
\newcommand{\cS}{\mathcal{S}}
\newcommand{\cP}{\mathcal{P}}
\newcommand{\Si}{\bm{\sigma}}
The DT of a binary mask, specifies the minimum distances from each pixel to the boundaries of non-zero regions, where the distances may be signed to distinguish between the inside and outside of the non-zero regions. Instead of the boundaries, unsigned DTs compute minimum distances to the whole masked regions. Various unsigned DTs have been studied for image segmentation in the literature~\cite{GeoS,DIOS,DeepIGeoS}. Given an N-Dimension gray image $I$ and a corresponding binary mask $M$, $I(\xx)$ represents the image intensity at point $\xx$ and $M(\xx)\in\{0, 1\}$. Point set $\cS$ is defined as $\{\xx\vert M(\xx)=1\}$ that is considered the point set of user interactions. The DT of $I$ concerning $\cS$ is formulated as:
\begin{equation}\label{eq:dist}
    D(\xx, \cS, I) = \min_{\xx'\in\cS}d(\xx, \xx', I),\quad \forall\xx\in I,
\end{equation}
where $d(\cdot,\cdot)$ is a specific distance function between two points in an image. For Euclidean distance and geodesic distance, $d(\cdot,\cdot)$ can be uniformly defined as:
\begin{equation}
    d(\xx, \yy, I) = \min_{\Ga\in\cP_{\xx,\yy}}\int_{0}^1\sqrt{\alpha\Vert\Ga'(s)\Vert^2 + \beta(\nabla I\cdot\uu)^2}\,ds,
\end{equation}
where $\cP_{\xx,\yy}$ is the set of all paths between the point $\xx$ and $\yy$, and $\Ga(s):\mathbb{R}\rightarrow\mathbb{R}^2$ is such a path, parameterized by $s\in[0, 1]$. $\Ga'(s)$ is the derivative of $\Ga(s)$ with respect to $s$ and $\uu=\Ga'/\Vert\Ga'\Vert$ is a unit vector along the tangent direction of $\Ga$. If $\beta=0$, Equation~(\ref{eq:dist}) is called Euclidean distance transform, which is not conditioned on the image intensity and thus degenerates to $D_{euc}(\xx,\cS)$. If $\alpha=0$, it becomes a geodesic distance transform. When $\alpha\neq0$ and $\beta\neq0$, Equation~(\ref{eq:dist}) is a combination of the two distances.  A common characteristic of the two distance functions is that they strongly rely on the image size, which means that for images with different sizes, the intensity distribution is significantly different. We name the DT with this feature as ``global transform''. An illustration of EDT, GDT and their blended DT is shown in Fig.~\ref{fig:dist}~(c-e). It can be seen that the grayscale (shown as the color bar) significantly varies from the image size (shown as the dotted box).

\begin{figure*}[!t]
\centering
\includegraphics[width=0.98\linewidth]{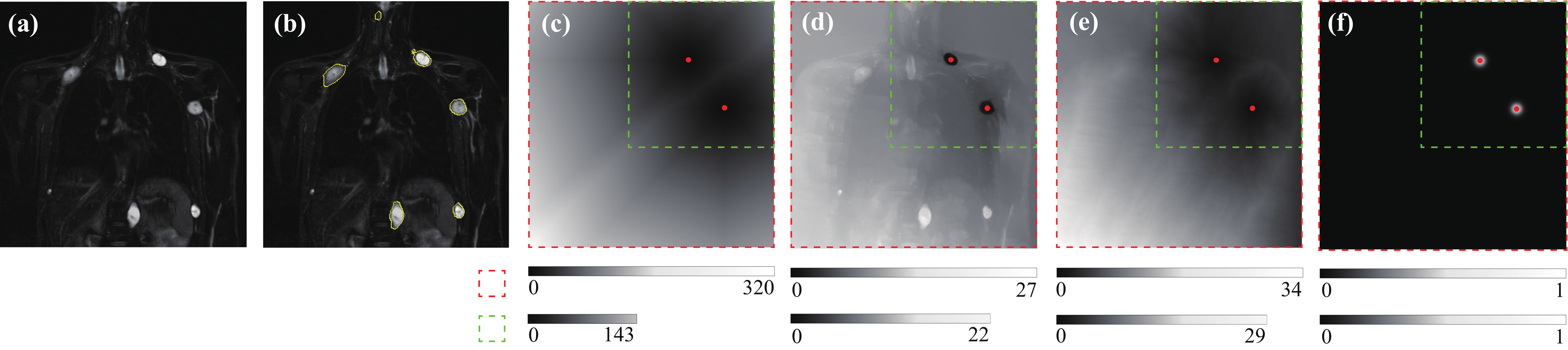}
\caption{Examples of different DTs. (a)~Raw image; (b)~Ground truth; (c)~Euclidean distance transform; (d)~Geodesic distance transform; (e)~Blend of Euclidean and geodesic distance; (f)~Exponential distance transform. Interactions are displayed as red points, i.e., the set $\cS$. Color bars represent the grayscale of distance maps with different image sizes.}
\label{fig:dist}
\end{figure*}

\begin{figure}[!t]
\centering
\includegraphics[width=\linewidth]{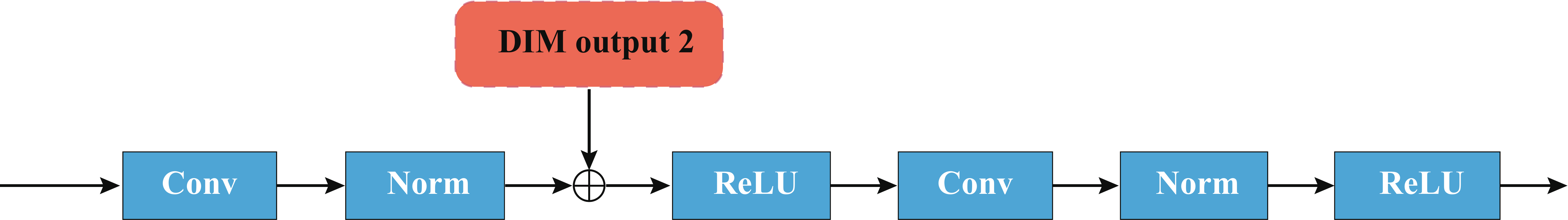}
\caption{Details of the deepest layer of the encoder. DIM output 2 are added to the output of the first normalization layer.}
\label{fig:dim-connect}
\end{figure}

Compared with popular CNN models taking fixed size images as input for classification problems~\cite{vgg,resnet}, FCN-like models, such as U-Net, remove densely connected layers and thus can accept input images of arbitrary sizes and produce correspondingly-sized outputs~\cite{fcn}. Furthermore, FCN-like models allow the size of inference images to be different from that of training images, which is crucial in the segmentation of large 3D medical images such as WBMRI due to limited GPU memory and insufficient training samples. For example, we may need to train a 3D U-Net with small volume patches due to the limited GPU memory, but inference with the whole volume as the inference saves the memory of storing gradients of parameters. Most of the operators in CNNs such as addition, multiplication, ReLU, convolution, and max pooling are either element-wise or window-wise~\cite{krizhevsky2012imagenet}, with which the predictions are hardly affected if the image patches were expanded or clipped (i.e., the size changes). However, the distribution of the integrated global DT corresponds to the actual feature size, and therefore inconsistent image sizes in training and inference stages lead to distribution inconsistency, which may impact the segmentation performance. Therefore, we propose a ``local transform'', exponential distance transform, to avoid being affected by the variable image sizes. 

The ExpDT is formulated as:
\begin{equation}
    ExpDT(\xx, \cS) = \max_{\xx'\in\cS}d_{exp}(\xx, \xx'),
\end{equation}
\begin{equation}\label{eq:expdt}
    d_{exp}(\xx, \yy) = \gamma\exp\left(-\frac{\Vert\xx-\yy\Vert_2^2}{2\Si^2}\right)
\end{equation}
with $\gamma$ the scale parameter, and $\Si$ controlling the influence of the points in $\cS$ on surrounding points. As shown in Fig.~\ref{fig:dist}~(f), the ExpDT is a local-enhanced distance map that means pixels with high gray levels are tightly gathered near the points in $\cS$, and therefore ExpDT is hardly affected by the variable image sizes.  When $\Si\rightarrow+\bm{\infty}$, ExpDT tends to form spikes at the point in $\cS$; and when $\Si\rightarrow\mathbf{0}$, ExpDT becomes flat and loses locality. Different from the previous two transforms that use {\it min} to compute distances to $\cS$, ExpDT turns to use {\it max} due to the negative sign in equation~(\ref{eq:expdt}). If we only considered from the perspective of DTs, ExpDT neither has global attributes nor combines image intensity. We argue that, with the proposed DINs framework, CNNs can still learn discriminative features from the local-enhanced ExpDT. 

\subsection{Structure of DINs}\label{sec:DIN}
The structure of DINs follows the encoder-decoder scheme with skip connection like 3D U-Net~\cite{3d-unet}. But 3D U-Net is originally evaluated to segment organs with fixed size and balance pixel spacing, such as Xenopus kidney, and not suitable for the scenario of various neurofibromas. Considering that WBMRI has an approximate average shape of $20\times1080\times321$ with a pixel spacing of $10\times1.56\times1.56mm$, we fix the size of the input image to $10\times512\times160$, which is about half the size of the original image. It should be noted that when the pixel spacing of an image differs significantly, isotropic resampling is not a good choice, due to the missing inter-slice information. Therefore, we build the backbone network with convolutional layers that have different kernel sizes and strides. There are four downsamplings in the coronal plane, but only once in the orthogonal direction. Instead of using max pooling layers, downsampling is implemented by large-stride convolution to save GPU memory and apply a larger batch size. Upsampling is performed by deconvolutional layers~\cite{deconv}. Commonly, batch normalization~(BN), which is used to reduce internal covariate shift and stabilize training~\cite{batch-norm}, has poor performance when confronted with small batch size~\cite{group-norm}. In addition, Isensee \etal~\cite{isensee2019nnu} experimentally demonstrate instance normalization~\cite{instance-norm}~(IN) performs better than BN in medical images. Therefore, we apply IN after the convolutional layers, followed by ReLU activation. For clarity, we list all the details of the internal layers of DINs in Table~\ref{tab:framework}.
 
\begin{table}[!t]
    \caption{\label{tab:framework}Details of the feature extractor of DINs. ``conv'' denotes a series of convolutional, instance normalization and ReLU layers and ``conv$^*$'' denotes pure convolutional layer. ``deconv'' means deconvolutional layer. And $[~]\times2$ denotes repeat the layer twice.}
    \centering
    \begin{tabular}{>{\centering}p{2cm}l}
    \hline
    Modules & Details of the layers \\ \hline
    Input   & 1 channel~$\leftarrow$ concat: DIM output 1 \\ \hline
    E1      & [conv~~k:~$133, 30$~~s:~$111$]$~\times~2$ \\ \hline
    E2      & \makecell[l]{conv~~k:~$133, 60$~~s:~$122$ \\ conv~~k:~$133, 60$~~s:~$111$} \\ \hline
    E3      & \makecell[l]{conv~~k:~$333, 120$~~s:~$122$ \\ conv~~k:~$333, 120$~~s:~$111$} \\ \hline
    E4      & \makecell[l]{conv~~k:~$333, 240$~~s:~$122$ \\ conv~~k:~$333, 240$~~s:~$111$} \\ \hline
    E5      & \makecell[l]{conv~~k:~$333, 320$~~s:~$222\leftarrow$ add: DIM output 2 \\ conv~~k:~$333, 320$~~s:~$111$} \\ \hline
    D4      & \makecell[l]{deconv~~k:~$222, 240$~~s:~$222\leftarrow$ concat: E4 output \\ {[}conv~~k:~$333, 240$~~s:~$111$]$~\times~2$} \\ \hline
    D3      & \makecell[l]{deconv~~k:~$122, 120$~~s:~$122\leftarrow$ concat: E3 output \\ {[}conv~~k:~$333, 120$~~s:~$111$]$~\times~2$} \\ \hline
    D2      & \makecell[l]{deconv~~k:~$122, 60$~~s:~$122\leftarrow$ concat: E2 output \\ {[}conv~~k:~$133, 60$~~s:~$111$]$~\times~2$} \\ \hline
    D1      & \makecell[l]{deconv~~k:~$122, 30$~~s:~$122\leftarrow$ concat: E1 output \\ {[}conv~~k:~$133, 30$~~s:~$111$]$~\times~2$} \\ \hline
    Output  & conv*~~k:~$111, 2$~~s:~$111$ softmax \\
    \hline
    \multicolumn{2}{l}{``k: 133, 30''--kernel size (1, 3, 3) and output channel 30} \\
    \multicolumn{2}{l}{``s: 111''--stride (1, 1, 1)}
    \end{tabular}
\end{table}
    
To incorporate user interactions into deep neural networks, we develop a deep interactive module (DIM). By leveraging feature modulation~\cite{film}, the DIM embeds additional image-specific information into the network backbone and guides the model to focus on the features that are enhanced by the distance maps, or the so-called guide maps, as shown in Fig.~\ref{fig:framework}. The DIM consists of an ExpDT, a max pooling layer, and a convolutional layer, transforming user interactions into guided maps of two different sizes (DIM output 1 and DIM output 2). Concretely, user interactions are transformed into a foreground guide map and a background guide map by ExpDT and are then integrated into the input layer by concatenating with the raw image as a three-channel input. The two guided maps are further encoded and integrated into the encoder's deepest layer to avoid the guide information being gradually diluted as more complex features are extracted. A detailed experiment regarding the position where the DIM outputs are inserted is placed in Section~\ref{sec:DIM}.  As shown in Fig.~\ref{fig:dim-connect}, downsampled guide maps are added to the output of the first normalization layer in the deepest layer of the encoder and followed by a ReLU activation function. The layers in decoder path do not need more integration due to the guide information passed from the skip connections. 

\subsection{Simulating strategy}\label{sec:simul}
\newcommand{\cF}{\mathcal{F}}
\newcommand{\cB}{\mathcal{B}}
\newcommand{\cO}{\mathcal{O}}
\newcommand{\cR}{\mathcal{R}}
Simulating user interactions in the training stage and evaluation stage can not only free users from the burdensome interactive work for generating thousands of training samples, but also accelerate the process of exploring the optimal hyperparameters. Our strategy of simulating user interactions is based on the work in~\cite{DIOS} and extends to the setting of 3D images. Let $M$ denote the ground truth segmentation of an image $I$, and $\cF$ denote the set of pixels of foreground objects satisfying $M(\xx)=1,\;\forall\xx\in\cF$. We define background regions surrounding objects as:
\begin{equation}
    \cB=\{\xx\in I\vert\xx\notin\cF, D_{euc}(\xx, \cF)<w\},
\end{equation}
where $D_{euc}(\xx, \cF)$ is the Euclidean distance between the point $\xx$ and the set $\cF$, and $w$ is the bandwidth. 

When processing 2D natural images, Xu \etal~\cite{DIOS} proposed to sample positive clicks from $\cF$ randomly, and the number of sampled points followed a discrete uniform distribution from 1 to $N_{pos}$. Negative clicks were randomly selected from the whole background~({\it random selection}) and evenly select from $\cB$~({\it uniform selection}). The number of negative points did not exceed $N_{neg}$ but could be zero. However, if we directly use the same upper bound $N_{pos}$ and $N_{neg}$ in 3D images, user interactions will become quite sparse as the additional axis, which is experimentally demonstrated to be harmful to the model performance. Therefore, we adapt $N_{pos}$ and $N_{neg}$ to the 3D version as follow: 
\begin{equation}\label{eq:num_inters}
    N_{pos,3D} = N_{pos}^{3/2}\qquad N_{neg,3D} = N_{neg}^{3/2}.
\end{equation}
This strategy maybe not the best choice, but it indeed is a simple and effective way to determine a better upper bound of the click numbers sampled in the training stage.
In addition, positive and negative clicks from the $d_{margin}$-pixel region near the boundaries should be avoided. Considering the fact of large inter-slice spacing of WBMRIs and infiltrative MRI appearance of neurofibromas, the restriction of $d_{margin}$ was only applied to individual slices. Finally, at least $d_z, d_y, d_x$ pixels should be kept between any two points in each dimension. 

During the evaluation, simulation is performed by placing the next positive/negative click on the center of the largest error region acquired from the symmetric difference between the current prediction and ground truth. Specifically, if the largest error region $\cR$ is part of a foreground object, then the next click $\xx$ is a positive point, whose coordinate is $\xx=1/\vert\cR\vert\sum_{\xx_i\in\cR}\xx_i$.
If $\xx\notin\cR$, we replace $\xx_f$ by:
\begin{equation}
    \newcommand{\argmin}{\mathop{\mathrm{arg\,min}}}
    \xx'=\argmin_{\yy\in \text{ske}(\cR)} D_{euc}(\xx, \yy)
\end{equation}
where $ske(\cR)$ is the skeleton~\cite{skeleton} of the region $\cR$. It means that $\xx'$ is the nearest point of $\xx$ in the skeleton of $\cR$. This situation may occur when $\cR$ is concave.
In this way, we guarantee that the positive points will not be placed in the wrong region~(background) and vice versa. The maximum number of clicks on a single study is limited to $n_{inters}$. we set a threshold DSC~(see Section~\ref{sec:detail}) of $t_{DSC}$ in cross-validation experiments. If the target threshold can not be achieved in $n_{inters}$ clicks, we will terminate the interaction of the current study.

\begin{table*}[!t]
\caption{\label{tab:dist}Cross-validation results of ExpDT and other methods on the training set with a threshold of 80\% DSC. FG: foreground points, BG: background points, Overall: total number of interactions.}
\centering
\begin{tabular}{c|ccc|cccccc}
\hline
Methods & $\alpha$ & $\beta$ & $\Si$ & DSC $\uparrow$ & VOE $\downarrow$ & ARVD $\downarrow$ & FG  & BG  & Overall $\downarrow$ \\ \hline
EDT~\cite{DIOS}       & 1.0      & 0.0     & -             & 0.62 & 0.52 & 0.73 & 7.6 & 7.7 & 15.3 \\
EDT-half                & 1.0      & 0.0     & -             & 0.68 & 0.46 & 0.46 & 7.6 & 6.6 & 14.2 \\
GDT~\cite{DeepIGeoS}  & 0.0      & 1.0     & -             & 0.72 & 0.41 & 0.49 & 5.9 & 5.7 & 11.6 \\
GDT-half                & 0.0      & 1.0     & -             & 0.70 & 0.43 & 0.51 & 5.4 & 6.0 & 11.4 \\ 
(EDT + GDT)-half     & 0.5      & 0.5     & -             & 0.70 & 0.43 & 0.55 & 5.6 & 6.4 & 12.0 \\ \hline
\multirow{3}{*}{ExpDT (ours)} & -  & -       & $(1, 5, 5)$   & 0.74 & 0.39 & {\bf 0.25} & 7.1 & 3.1 & 10.2 \\
                        & -        & -       & $(2, 6, 6)$   & 0.73 & 0.40 & 0.33 & 4.8 & 5.5 & 10.3 \\
                        & -        & -       & $(1.5, 6, 6)$ & {\bf 0.75} & {\bf 0.38} & 0.26 & 5.2 & 4.3 & {\bf 9.5} \\
\hline
\end{tabular}
\end{table*}

\begin{table*}[!t]
\caption{\label{tab:dist-test}Comparison of ExpDT with other methods on the WBMRI test set with at most 20 interactions. The test set is more challenging than the training set due to the distribution shift compared with the training set. FG: foreground points, BG: background points, Overall: total number of interactions. $p$ is the p-values of t-test between the results of ExpDT and other DTs.}
\centering
\begin{tabular}{c|ccc|ccccc|c}
\hline
Methods & $\alpha$ & $\beta$ & $\Si$ & DSC $\uparrow$ & VOE $\downarrow$ & ARVD $\downarrow$ & FG & BG & $p$ \\ \hline
EDT~\cite{DIOS}       & 1.0      & 0.0     & -             & 0.36 & 0.73 & 62.25 & 4.2  & 15.8 & \textless 0.01 \\
EDT-half                & 1.0      & 0.0     & -             & 0.38 & 0.73 & 22.92 & 5.0  & 15.0 & \textless 0.01 \\
GDT~\cite{DeepIGeoS}  & 0.0      & 1.0     & -             & 0.23 & 0.86 & 86.38 & 2.3  & 17.7 & \textless 0.01 \\
GDT-half                & 0.0      & 1.0     & -             & 0.53 & 0.60 & 10.53 & 6.0  & 14.0 & \textless 0.01 \\ 
(EDT + GDT)-half               & 0.5      & 0.5     & -             & 0.43 & 0.68 & 30.65 & 4.1 & 15.9 & \textless 0.01 \\ \hline
\multirow{3}{*}{ExpDT (ours)} & -  & -       & $(1, 5, 5)$   & 0.61 & 0.52 & {\bf 0.59}  & 10.3 & 9.7 & 0.10 \\
                        & -        & -       & $(2, 6, 6)$   & 0.66 & 0.47 & 0.91  & 7.2  & 12.8 & 0.33 \\
                        & -        & -       & $(1.5, 6, 6)$ & {\bf 0.67} & {\bf 0.46} & 0.79  & 7.6  & 12.4 & - \\
\hline
\end{tabular}
\end{table*}

%% file: subsections/4-Experiments.tex
\section{Experiments settings}

\subsection{Data set and preprocessing}\label{sec:data}
We collected two WBMRI data sets and an LRMRI data set from NF1 patients; one WBMRI data set was used as the training set and the remaining two as testing sets. Both WBMRI data sets were acquired on 1.5-T MR scanners (MAGETOM Avanto fit, Siemens Medical Systems, USA) using different software (Syngo MR 2004 V for the training set, Syngo MR E11 for the testing set). We did not shuffle and reassigned the training and testing sets to evaluate the DINs' ability to handle such a complicated situation. The training set contained 125 studies with 1156 NF1 tumors manually contoured by clinicians with expertise in identifying peripheral nerve sheath tumors. Their sizes and pixel spacings are described in section~\ref{sec:DIN}. 
We adopted online data augmentation to reduce overfitting, including randomly cropping from MRI scans, scaling between 1.0 and 1.25, rotating with a degree $z$ sampled from Gaussian distribution $\mathcal{N}(0, 5)$, flipping in both three dimensions, and gamma transformation with a range of $[0.7, 1.5]$.
The WBMRI testing set was composed of 33 studies with an approximate dimension of $41\times1340\times397$ and unified spacing of $6\times1.30\times1.30mm$. A total of 224 tumors were manually contoured in this data set. The LRMRI testing set contained 45 studies with various dimensions (from $24\times224\times224$ to $149\times768\times768$) and spacing (from $3.93\times0.42\times0.42mm$ to $10\times1.33\times1.33mm$).

\subsection{Implementation details}\label{sec:detail}
We used the weighted cross-entropy as the loss function. The weighting factors were 1.0 and 3.0 for background pixels and foreground pixels, respectively. Adam optimizer~\cite{adam} with $\beta_1=0.9$ and $\beta_2=0.99$ was used to update model parameters. The learning rate was set to $0.0003$ at first and reduced by 0.2 once the validation loss did not decrease for 30 epochs. The minimum learning rate was set to $1e^{-6}$. We trained 200 batches per epoch with a batch size of 8 and terminated training after 250 epochs. We forced 50\% of the images in one batch to include tumors while the others were randomly cropped without restriction. We implemented DINs with the Tensorflow~\cite{tensorflow} package in Python, and conducted experiments on a single Tesla V100 GPU with 32GB memory. We utilized ITK-SNAP~\cite{py06nimg}, 3DQI~\cite{CAI2018144} and imcut~\cite{jirik2013} for comparing with Active Contour, Random Walk, and Graph Cut, respectively.

Foreground and background clicks were simulated following the strategies described in Section~\ref{sec:simul}. In DIOS, $N_{pos}$ is set to $5$ in 2D image segmentation. We therefore set
\begin{equation}
N_{3D}:=N_{neg,3D}=N_{pos,3D}=5^{3/2}\approx 11    
\end{equation}
The maximum number of background clicks $N_{neg,3D}$ was set to the same value as $N_{pos,3D}$ for simplicity. It should be noted that this is a crucial hyper-parameter for acceptable performance, and we will discuss it in section~\ref{sec:ablation}. 
$d_{margin}$ was set to 3 pixels by default. To preserve tumors less than 6 pixels in either direction, we remove the restriction of $d_{margin}$ for these small tumors. $d_y$ and $d_x$ were set to 10, and $d_z$ was liberalized to 1 as a compromise of the huge inter-slice spacing. The bandwidth of $\cB$ was set to 40 pixels. 

In the evaluation, dice similarity coefficient~(DSC), the total number of clicks, Volumetric overlap error~(VOE), and absolute relative volume difference~(ARVD) were the primary evaluation metrics. The number of positive clicks and negative clicks were also logged for comparison. Let $P$ and $Q$ denote the binary prediction and ground truth, respectively. Then, the three metrics are formulated as:
\begin{equation}
    DSC(P, Q) = \frac{2\vert P\cap Q\vert}{\vert P\vert + \vert Q\vert},
\end{equation}
\begin{equation}
    VOE(P, Q) = 1 - \frac{\vert P\cap Q\vert}{\vert P\cup Q\vert},
\end{equation}
\begin{equation}
    ARVD(P, Q) = \frac{abs(\vert P\vert - \vert Q\vert)}{\vert Q\vert},
\end{equation}
where $\vert\cdot\vert$ denotes the number of non-zero elements and $abs$ means the absolute value. 
For the interactive evaluation process, if not specified, user interactions were continuously provided until either of 20 clicks or the threshold of 0.8 DSC was reached in all of the following experiments.

%% file: subsections/5-Results.tex
\section{Results and Discussion}

\subsection{Comparison of ExpDT with EDT and GDT}
In this section, we compare ExpDT with EDT, GDT, and other three variants. We remove the DIM output 2 in this group of experiments to highlight the contributions of DT. The suffix ``half'' denotes the input images downsampled to half of the original image patches. The cross-validation results are listed in Table~\ref{tab:dist}. When using settings of $\Si=(1,5,5)$, ExpDT outperforms EDT by 12\% DSC and reduces the average number of interactions from 15.3 to 10.2. Compared with GDT, ExpDT does not involve image intensity but achieves better results in all metrics. Interestingly, with a DSC threshold of 80\%, EDT-half and GDT-half achieve fewer interactions than EDT and GDT, respectively. We conjecture that the smaller image sizes alleviated the problem of inconsistency of image sizes, which affects the performance of ``global transformation'', between the training and evaluation stages. Finally, ExpDT with $\Si=(1.5,6,6)$ achieves the best results on all the four primary metrics.
The last three rows of Table~\ref{tab:dist} indicate that a successive improvement can be made by subtly adjusting $\Si$ even if the model parameters are fixed after training. 
The comparison between ExpDT with other DTs demonstrates the effectiveness and flexibility of ExpDT. 

\begin{figure}[!t]
\centering
\includegraphics[width=0.95\linewidth]{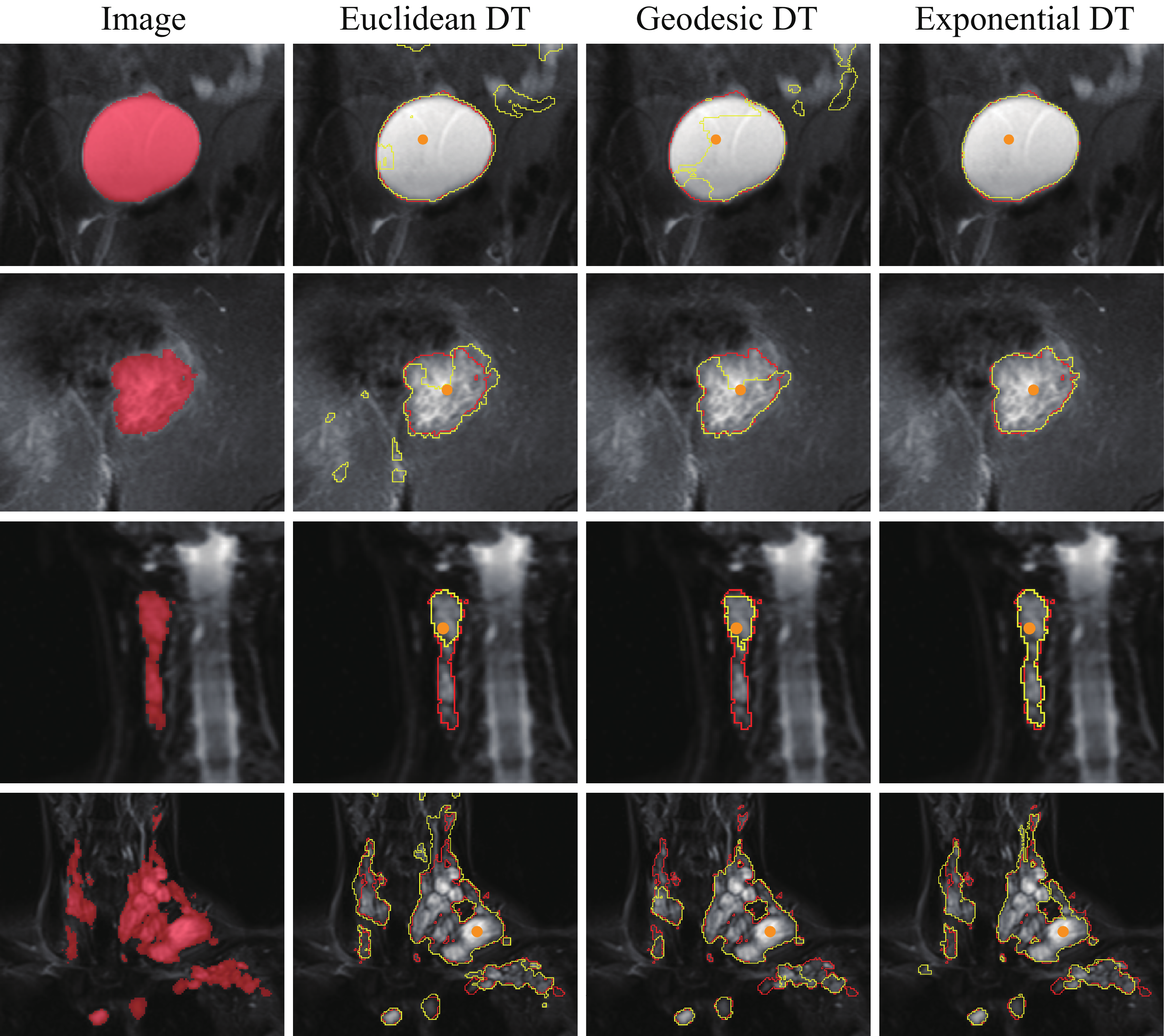}
\caption{Examples of the segmentation results by different DTs. Each row is a coronal patch from the training set. The first column shows the raw images and the corresponding ground truths. The right three columns are the segmentation results of three DTs: EDT, GDT, and ExpDT, respectively. Red curve: the boundaries of the ground truth. Yellow curve: the boundaries of segmentation results.}
\label{fig:res-display}
\end{figure}

\begin{figure}[!t]
\centering
\includegraphics[width=0.93\linewidth,height=0.38\textwidth]{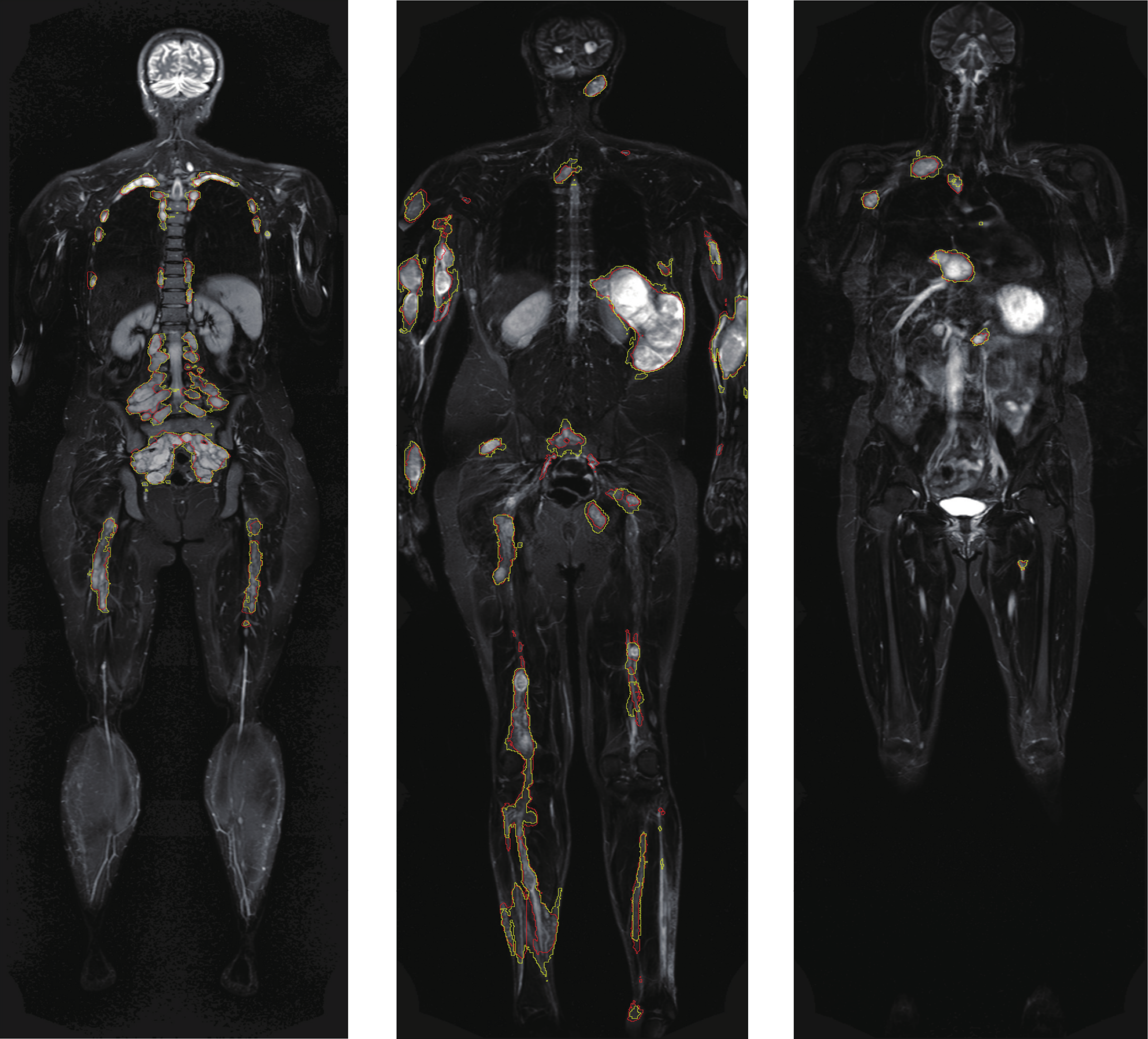}
\caption{Examples of the segmentation results on WBMRI by DINs. Red curve: the boundaries of the ground truth. Yellow curve: the boundaries of segmentation results.}
\label{fig:wbmri-display}
\end{figure}

Table~\ref{tab:dist-test} compares the performance of different DTs on the more challenging testing set. We report three accuracy metrics and the numbers of foreground and background points with 20 clicks provided. One can observe that the overall performance is inferior to those in the training set. Potential reasons include the difference in image size, pixel spacing, and the distribution shift between training and test set. We observe that the accuracy of EDT and GDT decreases more than ExpDT. The number of background points is much more than the number of foreground points, and the ARVD is notably larger than that of ExpDT. It indicates EDT and GDT predict many false-positive regions on account of the properties of their ``global transformation''. On the contrary, ExpDT yields improved results, and the numbers of positive clicks and negative clicks are relatively balanced. We also report the p-values of t-test between ExpDT and other DTs, which indicate the substantial improvement of ExpDT. Therefore, as a ``local transformation'', ExpDT is more generalizable.

Fig.~\ref{fig:res-display} displays four segmentation cases by DINs with different transform functions. For each case, only one positive click is provided to perform segmentation. We observe that with only one click, DINs with ExpDT achieved better performance in the segmentation of discrete neurofibromas~(first row and second row), whereas EDT and GDT produce more false-positive regions as well as false-negative regions. In plexiform neurofibromas, ExpDT achieves consistent performance to segment with good accuracy~(third row and fourth row). However, EDT and GDT may miss tumor regions far from the clicked object~(third row). Their high sensitivity to image size is the main reason for such instability. For large plexiform neurofibromas (fourth row), GDT may miss part of the lesions due to heterogeneous and diffuse tumor architecture, while ExpDT and show better performance. Fig.~\ref{fig:wbmri-display} shows some results on WBMRI by DINs.

\begin{table}[!t]
\caption{\label{tab:mgh-comp}Comparison of DINs with other neurofibroma segmentation methods on the LRMRI test set (45 cases). VD: Volume difference, $\vert P\vert-\vert Q\vert$; RVD: Relative volume difference, $2\vert P-Q\vert/(\vert P\vert+\vert Q\vert)$. The values in VD and RVD columns are median (range). Num: Number of cases with <20\% RVD.}
\centering
\begin{tabular}{c|ccc}
\hline
Methods & VD $\downarrow$ & RVD (\%) $\downarrow$ & Num. $\uparrow$ \\ \hline
NCI-3DQI~\cite{CAI2018144} & -21 (-406 to 114) & 4.5 (0.3 to 28.4) & 43 \\
MGH-3DQI~\cite{CAI2018144} & -30 (-782 to 353) & 9.5 (0.1 to 48.8) & 34 \\
DINs                       & -3 (-101 to 360) & 16.6 (7.9 to 29.2) & 40 \\
\hline
\end{tabular}
\end{table}

\begin{table}[!t]
\caption{\label{tab:deep-cnn-comp}Comparison of DINs with deep CNN-based methods on the WBMRI test set with 20 interactions. ``A'' and ``I'' means automated and interactive methods, respectively. $p$ is the p-values of t-test between DINs and other methods.}
\centering
\setlength\tabcolsep{4pt}
\begin{tabular}{c|c|ccc|c}
\hline
Type & Methods & DSC $\uparrow$ & VOE $\downarrow$ & ARVD $\downarrow$ & $p$ \\ \hline
\multirow{3}{*}{A} & DeepMedic~\cite{kamnitsas2017efficient} & 0.06 & 0.97 & 77.95 & \textless 0.01 \\
& 3D U-Net~\cite{3d-unet}                 & 0.06 & 0.97 & 77.62 & \textless 0.01 \\
& nnU-Net~\cite{nnunet}                   & 0.25 & 0.84 & 5.98 & \textless 0.01 \\ \hline
\multirow{3}{*}{I} & nnU-Net + DIM output 1 & 0.40 & 0.72 & 6.70 & \textless 0.01 \\
& DINs (DIM output 1)                     & 0.67 & 0.46 & 0.79 & 0.58 \\
& DINs                  & {\bf 0.69} & {\bf 0.45} & {\bf 0.64} & - \\
\hline
\end{tabular}
\end{table}

\begin{figure*}[!t]
\centering
\includegraphics[width=\linewidth]{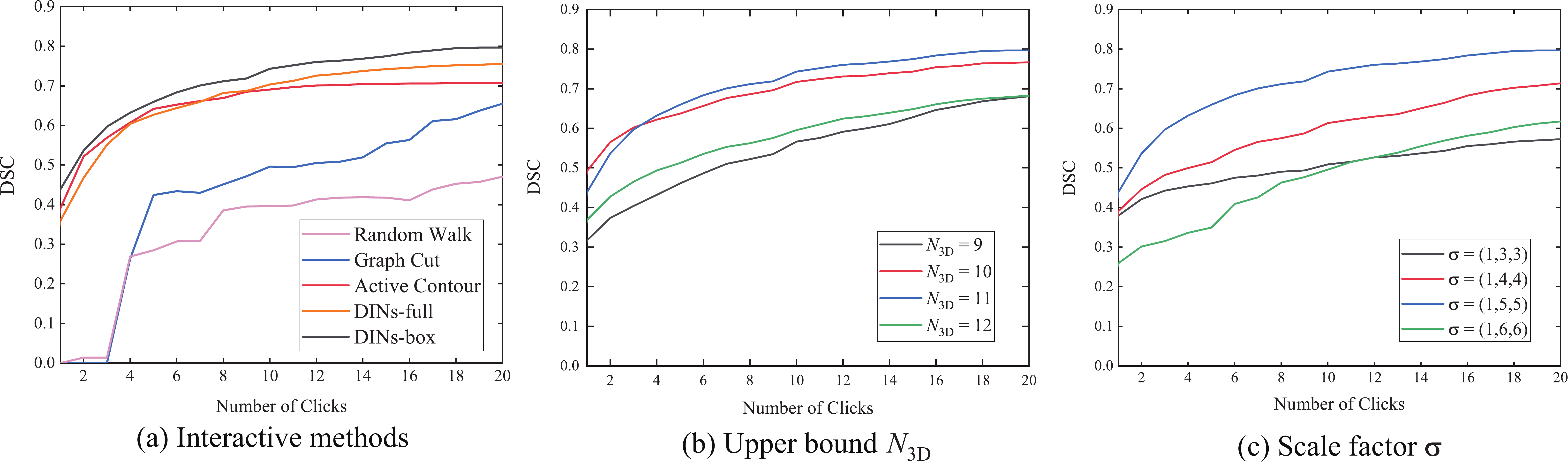}
\caption{Average DSC vs. the number of interactions on the WBMRI test set. Comparison of (a)~different interactive methods (b)~upper bound $N_{3D}$ of the number of interactions during training and (c)~scale factor $\Si$ of ExpDT.}
\label{fig:trend}
\end{figure*}

\begin{table*}[!t]
    \caption{\label{tab:inter}Average Number of interactions of the proposed methods with conventional interactive methods on the WBMRI test set with a threshold of 0.8 DSC. RW: Random Walk; GC: Graph Cut; AC: Active Contour; inters: interactions. The number of interactions count only the points or bubbles.}
    \centering
    \begin{tabular}{ccccc}
        \hline
        Methods                & Type of interactions & DSC (20 inters) $\uparrow$ &\# of inters (0.8 DSC) $\downarrow$ & Running time/inter (in minute) $\downarrow$ \\ \hline
        RW~\cite{randomwalk} & boxes, points & 0.47 & 19.5 & 2.12 \\
        GC~\cite{graphcut}   & boxes, points & 0.66 & 17.9 & 0.71 \\
        AC~\cite{snakes}     & boxes, thresholds, bubbles & 0.71 & 15.2 & 5.32 \\ \hline
        DINs-full (ours)       & points                 & 0.76 & 14.6 & 0.33\\
        DINs-box (ours)        & boxes, points & {\bf 0.80} & {\bf 12.1} & {\bf 0.14} \\
        \hline
    \end{tabular}
\end{table*}

\subsection{Comparison with other methods}

\subsubsection{State-of-the-art interactive methods}
Cai \etal~\cite{CAI2018144} perform volume measurements on the LRMRI data set using 3DQI software at Massachusetts General Hospital (MGH) and National Cancer Institute (NCI) and MEDx software at NCI.
Table~\ref{tab:mgh-comp} shows the results of NCI-3DQI - NCI-MEDx and MGH-3DQI - NCI-MEDx at the first two rows. We take the segmentation results of NCI-MEDx as the ground truth and use DINs on the LRMRI data set. Results in the last row indicate that DINs can achieve similar performance comparing with 3DQI. Notice that the results of NCI-3DQI and MGH-3DQI are finalized with various editing tools, while DINs are not for a fair comparison.

\subsubsection{Deep CNN-based methods}
A comparison between DINs and state-of-the-art medical image segmentation methods is shown in Table~\ref{tab:deep-cnn-comp}. One can observe that DINs outperform automated methods by 44\%--63\% and outperform ``nnU-Net + Dim output 1'' by 29\%. Automated methods get low scores, while interactive methods have better performance with the help of user knowledge. The comparison between ``nnU-Net + Dim output 1'' and ``DINs (DIM output 1)'' indicates that DINs benefit from the proposed structure of feature extractor adapted to WBMRI. Finally, with the DIM output 2, DINs further increase the DSC by 2\%, which suggests the effectiveness of integrating user knowledge into deeper layers. In addition, the p-values of t-test between the results of DINs and other methods are listed for reference.

\subsubsection{Conventional interactive methods}
We compare DINs with some conventional interactive segmentation methods, including Random Walk~(RW)~\cite{randomwalk}, Graph Cut~(GC)~\cite{graphcut} and Active Contour~(AC)~\cite{snakes}. RW and GC treat the volume as a discrete static graph, performing segmentation with many positive and negative clicks by solving the linear system~(RW) or min-cut~(GC) problem. Commonly, to save computation time and reduce irrelevant information, a bounding box is provided before running these conventional approaches. Then the search space can be restricted to a smaller region. Therefore, we implement two versions of DINs:
\begin{itemize}
    \item {\it DINs-full.} Feeding the whole 3D volume into DINs for evaluation.
    \item {\it DINs-box.} We manually create several bounding boxes in each volume with three criteria: (1)~Spatially closer tumors are grouped into the same bounding box. (2) The heights and the widths of bounding boxes are at least 128 pixels, while depths are set as tight to the tumor boundaries as possible that is consistent with users' behavior. (3) There are no more than five bounding boxes in one case.
\end{itemize}

\begin{figure*}[!t]
\centering
\includegraphics[width=\linewidth]{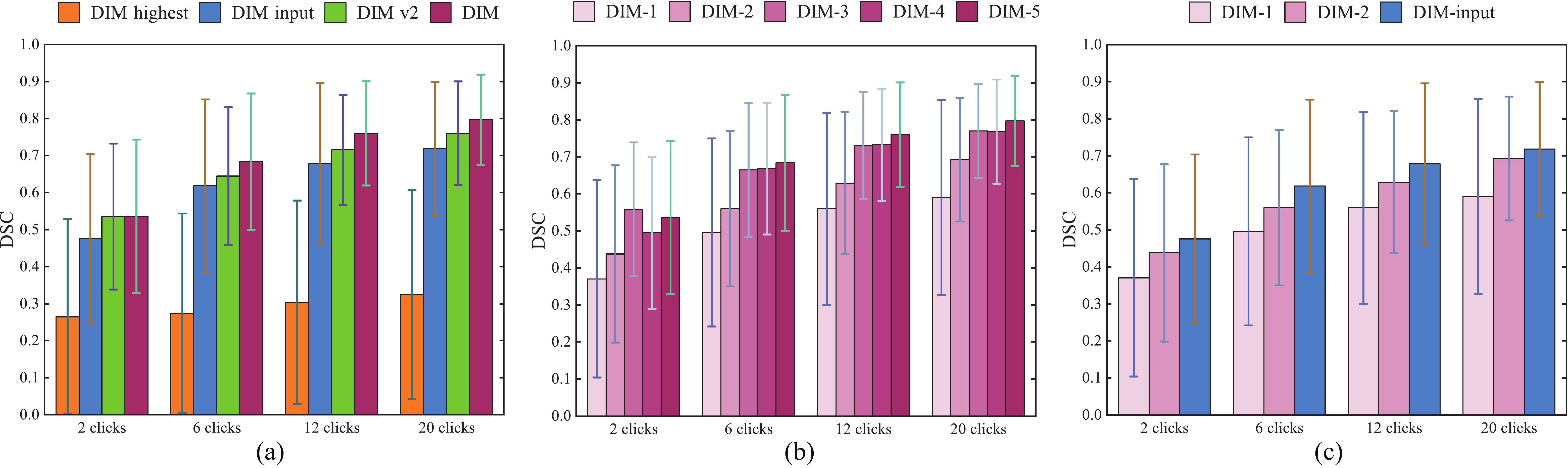}
\caption{Ablation study of the structure of DIM. The models are evaluated on the WBMRI test set. Average DSCs and standard deviations after 2, 12, 6, 20 clicks are presented in the bar chart. (a)~Comparison of the DIM components. (b)~Comparison of the effects of the position where the DIM output 2 is inserted in the encoder. (c)~A further comparison of DIM-2, DIM-3, and DIM-input.}
\label{fig:DIM-cmp}
\end{figure*}

\begin{figure*}[!t]
\centering
\includegraphics[width=\linewidth]{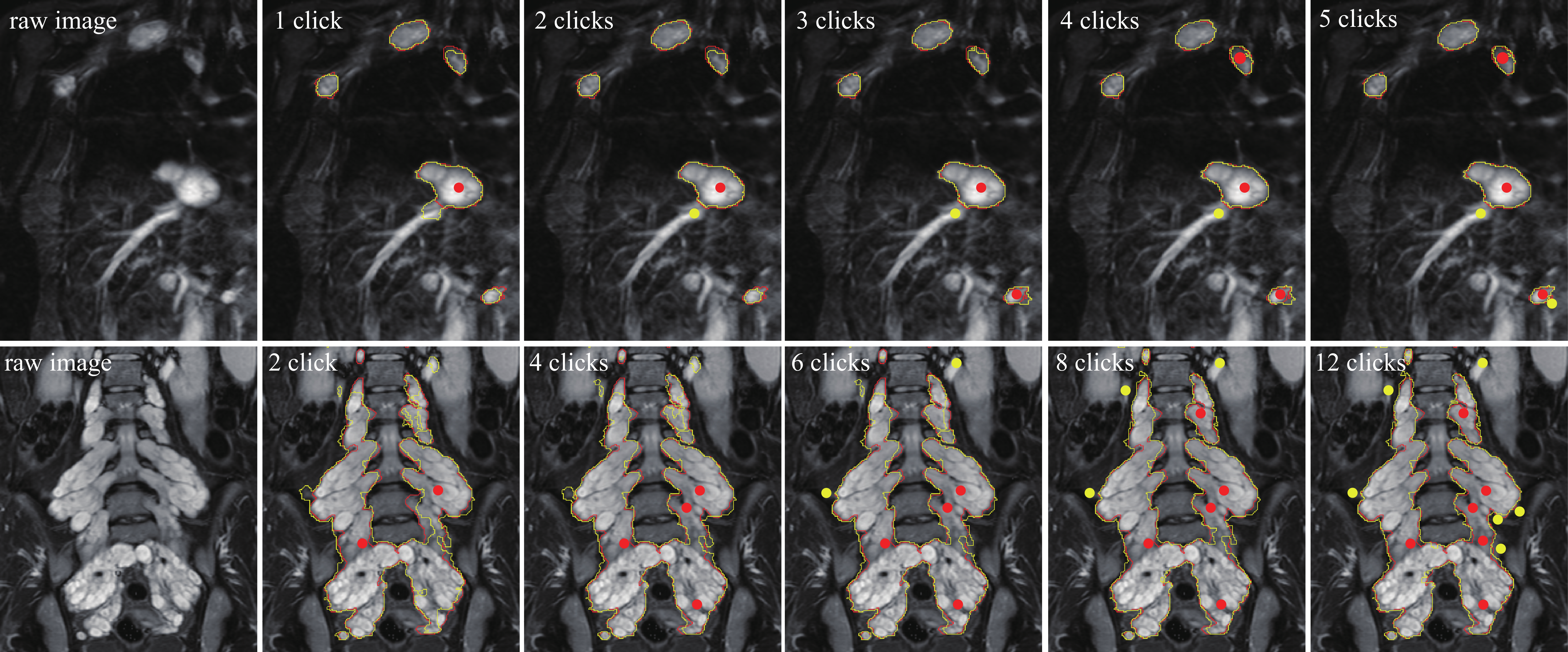}
\caption{Interactive segmentation results of plexiform neurofibromas with the DINs framework. Bounding boxes are used to trim noises and improve inference speed. The ground truth segmentation is shown in red contours and the prediction contour is shown in yellow. The positive and negative clicks are masked by red points and yellow points, respectively. }
\label{fig:inter-display}
\end{figure*}

Fig.~\ref{fig:trend}~(a) presents the trend of DSC for DINs and conventional methods when the interactive points are continuously provided. RW, GC, AC, and DINs-box ran the algorithms within the volume of interest while DINs-full employs the entire 3D volume.  In Fig.~\ref{fig:trend}, we observe that RW and GC show a low DSC, which is caused by the limited information used to compute features. We notice that DINs-full only has a modest improvement compared with AC since AC needs to set a bounding box and adjust two thresholds for filtering backgrounds, which requires much more user efforts to precisely tune the thresholds. With additional bounding boxes, DINs-box significantly exceeds all of the other three conventional interactive methods. For a detailed comparison, we summarize the type of interactions, DSC, number of interactions, and running time in Table~\ref{tab:inter}. DINs-full has the minimum requirement of interactions to perform the segmentation, which substantially reduces the complexity of the interaction. With extra bounding boxes, DINs-box further improves segmentation accuracy and reduces both the user burden and the running time. 

Overall, as the number of interactions increases, DINs improve the segmentation accuracy more stably and successively. Furthermore, a substantial improvement can be made by providing a few bounding boxes for some difficult cases~(DINs-box). This implies that DINs are more effective and flexible for interactive segmentation.

\begin{figure*}[!t]
    \centering
    \includegraphics[width=\linewidth]{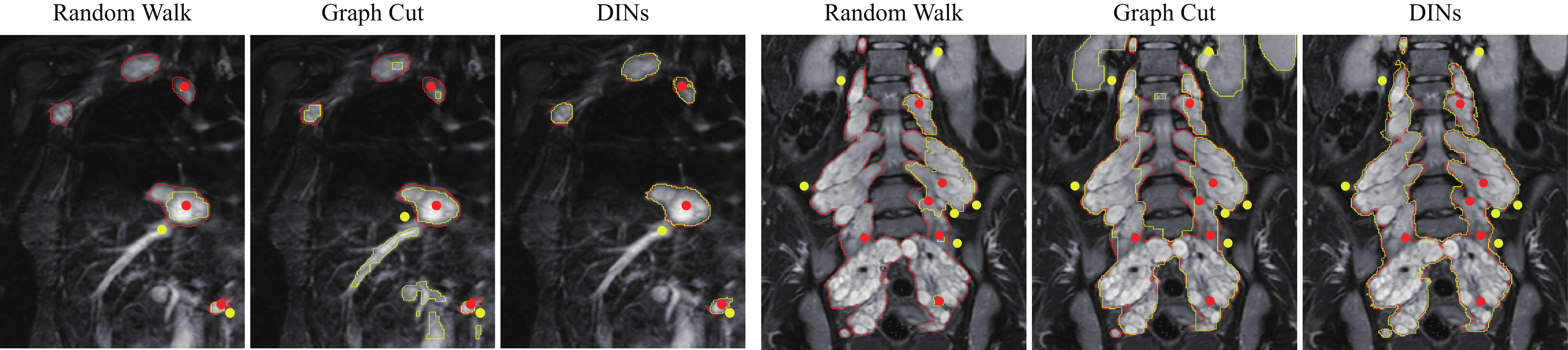}
    \caption{Comparison of DINs with GC and RW methods. The ground truth and the prediction ares shown in red and yellow contours, respectively. The positive and negative clicks are masked by red and yellow points, respectively. (A negative click in the second image has a deviated location compared with the other two methods due to the empty prediction with  original location.)}
    \label{fig:inter-cmp-display}
    \end{figure*}

\subsection{Ablation studies}\label{sec:ablation}

In this section, we conduct ablation studies to investigate the influence of three crucial parameters of DINs: (1)~the upper bound of the number of interactions sampled during training, (2)~scale parameter $\Si$ of ExpDT, and (3)~the DIM module.

\subsubsection{Number of interactions during training}
Training DINs with different upper bound $N_{3D}$, the number of foreground clicks and background clicks may significantly influence the performance of the resulting model. We conduct experiments with four different $N_{3D}\in\{\;9,10,11,12\;\}$ to train DINs and evaluate it on the test set. The comparison is shown in Fig.~\ref{fig:trend}~(b). We find that the model trained with $N_{3D}=11$ has a higher DSC than $N_{3D}=10$ when the number of clicks exceeds 3. The models trained with $N_{3D}=9$ and $12$ are significantly inferior to $N_{3D}=11$. Too many points in training may lead the model to severely rely on user interactions and become more conservative, while too few points may be not adequate to help the model choose discriminative features. It indicates that this hyper-parameter is crucial to train a well-performed model, and the strategy of linking the upper bound of interaction number and image dimension~(see Equation~(\ref{eq:num_inters})) is reasonable and effective.

\subsubsection{Scale parameter of ExpDT}
The scale factor $\Si=(\sigma_z, \sigma_y, \sigma_x)$ is another key parameter to train a well-performed model, where $z, y, x$ are correlated to the anatomical anterior-posterior, superior-inferior and left-right, respectively. We fix $\sigma_z$ to 1 and compared different scale values of $\sigma_x$ and $\sigma_y$. The results are presented in Fig.~\ref{fig:trend}~(c), which indicate that scale factor is a important parameter for optimal model performance. The potential reason is the varied sizes of neurofibromas. However, the sensitivity can also be seen as the flexibility that users can use different scale factors for tumors with various sizes to achieve higher accuracy. The scale factor can also be adjusted in the inference stage for better segmentation results. A comparison is shown in Table~\ref{tab:dist} and Table~\ref{tab:dist-test}. It indicates that a slightly larger $\Si$ gives better results. Furthermore, we can adjust the $\Si$ for each case to improve performance. 

\subsubsection{Deep interactive module}\label{sec:DIM}
We conduct experiments to present the effectiveness of each part of the DIM. Several variants of the DIM are compared: (1) DIM-input: DIM with only the output 1 branch. (2) DIM-highest: DIM with only the output 2 branch. (3) DIM-v2: The path to output 2 is implemented with a max-pooling layer whose kernel size is $(1,4,4)$ and two large-stride convolutional layers. The results are shown in Fig.~\ref{fig:DIM-cmp}~(a). Intuitively, the user interactions are additional features to discriminate against the tumor from the background. We observe that DIM-highest gets a poor DSC and only has subtle improvement as the number of clicks increase, while DIM-input has a higher DSC. By combining DIM-highest with DIM-input together, we achieve further improvement. The comparison among DIM-highest, DIM-input, and DIM indicates that guide maps with spatial information help neural networks learn more discriminative features, and the guide maps in the highest layer of encoder exactly enhance the corresponding features. A comparison between DIM and DIM-v2 indicates that a single convolutional layer is adequate to pass the extra features to deeper layers. More layers introduce more parameters that increase the risk of overfitting. 

We also compare the effect of the position where the DIM output 2 is inserted in the encoder. Let DIM-$n$ denote the variant of connecting DIM output 2 to the $n$th layer of the encoder, where $n\in\{1, 2, 3, 4, 5\}$. DIM-$5$ is exactly the proposed structure. The results are shown in Fig.~\ref{fig:DIM-cmp}~(b). Overall, the DSC increases as the guide maps are integrated into deeper layers. The reason is that guide maps' information is limited compared with the abundant features from the images and is easily diluted as more features are extracted. Therefore, it is the best choice to integrate the guide maps into the encoder's deepest layer for enhanced learning of the features. Besides, DIM-input outperforms DIM-1 and DIM-2, which is presented in Fig.~\ref{fig:DIM-cmp}~(c) for clarity. It indicates that integrating guide maps into shallow layers multiple times~(including the input layer) of the encoder hurts the feature learning because of overfitting, which impacts generalization.

\subsubsection{Effect of click positions}

Click positions will affect the interactive segmentation accuracy. To show the effect of click positions when using DINs, we randomly select ten neurofibromas from the training set, and each neurofibroma is clicked once. Each example is evaluated five times with different click positions. The standard deviations of the five results in each neurofibroma are computed. The median (range) of the ten standard deviations is 0.015 (0.005 to 0.044), which indicates that the click positions affect the performance within a reasonable range. As the click number increases to 2 and 3, segmentation results become more stable, and the median of the standard deviations decrease to 0.008 (0.001 to 0.038) and 0.005 (0.001 to 0.026), respectively.


\subsection{Interactive results}

Two interactive segmentation results of plexiform neurofibromas with DINs are displayed in Fig.~\ref{fig:inter-display}. The ground truth contours (manual segmentation) are red, and the prediction contours are yellow. The positive and negative interactive clicks are marked by red and yellow points, respectively. DINs achieve accurate segmentation of multiple tumors with one click and iteratively improves segmentation with additional interactions. Notice that the $\Si$ is set to $(1,5,5)$ by default, which is suitable for most neurofibroma segmentation situations.

In Fig.~\ref{fig:inter-cmp-display}, we compare three interactive methods with the same clicks. (Note: A negative click in the second image has a deviated location compared with the other two methods due to the empty prediction with the original location.) Random walk tends to lead to undersegmentation, while graph cut cannot distinguish neurofibromas from normal organs and tends to result in oversegmentation. In comparison, DINs can recognize neurofibromas accurately. The two groups of segmentation results support the advantages of the DINs.

%% file: subsections/6-Conclusion.tex
\section{Conclusion}

In conclusion, we propose the effective and flexible Deep Interactive Networks with a novel Exponential Distance Transform for neurofibroma segmentation on WBMRIs. The DINs framework efficiently extracts discriminative features of tumors by incorporating user interactions into low-level features and high-level features. The ``local transformation'' ExpDT is better equipped to address biased data distribution in medical images. Experiments on the training set and the test set show that the proposed method outperforms conventional interactive methods and performs significantly better than automated and interactive CNN-based methods. Limitations of DINs include the following parts: (1) Like conventional semi-automatic segmentation methods, DINs also need extra editing tools to achieve a acceptable volume measurement; (2) The ExpDT generates guide maps ignoring the image intensities, which may help improve the quality of the guide maps; (3) DINs may failed in some cases such as neurofibromas near the orbits due to the similar intensity. Considering these limitations, integrating the anatomical structure into neural networks and combining the image intensity into guide maps may be the future directions for developing high-performance interactive neural network methods.